\pdfoutput=1

\documentclass[11pt]{article}

\usepackage[final]{naacl2021}

\usepackage{times}
\usepackage{latexsym}

\usepackage{soul}
\usepackage{url}
\usepackage{xurl}
\usepackage{hyperref}
\usepackage{graphicx}
\usepackage{amsmath}
\usepackage{booktabs}
\usepackage{tabularx}
\usepackage{algorithm}
\usepackage{array}
\usepackage{algorithmic}
\usepackage{adjustbox}
\usepackage{diagbox}

\usepackage{xr}
\usepackage{CJKutf8}
\usepackage{mathtools, cuted}
\usepackage{amsmath,amsfonts,amssymb}
\usepackage{amssymb}
\usepackage{makecell}
\usepackage{stfloats}
\usepackage{multirow}
\usepackage{mathrsfs}
\usepackage{subfigure}
\usepackage{color}
\usepackage{listings}
\usepackage{enumitem}
\usepackage{booktabs}
\usepackage{framed}
\usepackage{caption}
\usepackage{natbib}
\usepackage{csquotes}
\usepackage[normalem]{ulem}
\usepackage[multiple]{footmisc}
\usepackage{kotex}

\newcommand{\chapternote}[1]{{%
  \let\thempfn\relax
  \footnotetext[0]{#1}
}}

\DeclareMathOperator*{\argmin}{arg\,min}

\usepackage[T1]{fontenc}

\usepackage[utf8]{inputenc}

\usepackage{microtype}
\title{Restoring and Mining the Records of the Joseon Dynasty via Neural Language Modeling and Machine Translation}

\author{Kyeongpil Kang \\
  Scatter Lab \\ Seoul, South Korea \\
  \href{kyeongpil@scatterlab.co.kr}{kyeongpil@scatterlab.co.kr} \\\And
  
  Kyohoon Jin \\
  Chung-Ang University \\ Seoul, South Korea \\
  \href{fhzh123@cau.ac.kr}{fhzh123@cau.ac.kr} \\\And
  
  Soyoung Yang \\
  KAIST \\ Daejeon, South Korea \\
  \href{sy_yang@kaist.ac.kr}{sy\_yang@kaist.ac.kr} \\\AND
  
  Soojin Jang \\
  Chung-Ang University \\ Seoul, South Korea \\
  \href{sujin0110@cau.ac.kr}{sujin0110@cau.ac.kr} \\\And
  
  Jaegul Choo \\
  KAIST \\ Daejeon, South Korea \\
  \href{jchoo@kaist.ac.kr}{jchoo@kaist.ac.kr} \\\And
  
  Youngbin Kim \\
  Chung-Ang University \\ Seoul, South Korea \\
  \href{ybkim85@cau.ac.kr}{ybkim85@cau.ac.kr} \\}

\begin{document}
\maketitle

\begin{abstract}
Understanding voluminous historical records provides clues on the past in various aspects, such as social and political issues and even natural science facts. However, it is generally difficult to fully utilize the historical records, since most of the documents are not written in a modern language and part of the contents are damaged over time. As a result, restoring the damaged or unrecognizable parts as well as translating the records into modern languages are crucial tasks. In response, we present a multi-task learning approach to restore and translate historical documents based on a self-attention mechanism, specifically utilizing two Korean historical records, ones of the most voluminous historical records in the world. Experimental results show that our approach significantly improves the accuracy of the translation task than baselines without multi-task learning. In addition, we present an in-depth exploratory analysis on our translated results via topic modeling, uncovering several significant historical events.
\end{abstract}

\section{Introduction}

\begin{figure*}[t]
\centering
\includegraphics[width=\textwidth,trim=0 0 0 0,clip]{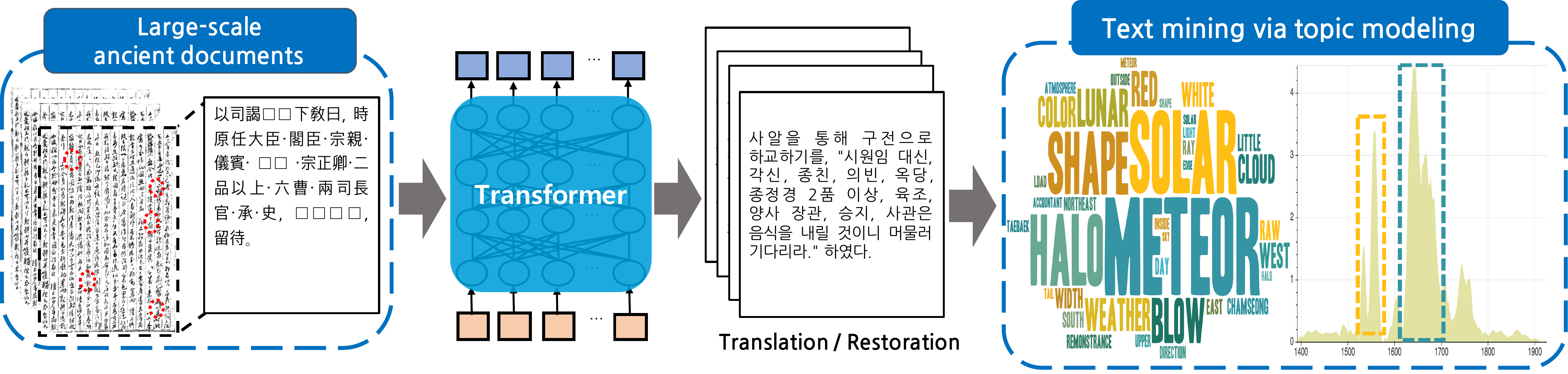}
\caption{Overview of the proposed approach of recovering, translating, and mining historical documents.}
\label{fig:overview}
\end{figure*}

Historical records are invaluable sources of information on the lifestyle and scientific records of our ancestors. Humankind has learned how to handle social and political problems by learning from the past. The historical records also serve as the evidence of intellectual accomplishment of humanity over time. 
Given such importance, there has been a great deal of nationwide efforts to preserve these historical records. For instance, UNESCO protects world heritage sites, and experts from all around the world have been converting and restoring historical records in a digital form for long-term preservation. A representative example is the Google Books Library Project\footnote{\url{https://support.google.com/websearch/answer/9690276}}.
However, despite the importance of the historical records, it has been challenging to properly utilize the records for the following reasons. First, the nontrivial amounts of the documents are partially damaged and unrecognizable due to unfortunate historical events or environments, such as wars and disasters, as well as the weak durability of paper documents. These factors result in difficulties to translate and understand the records. Second, as most of the records are written in ancient and outdated languages, non-experts are difficult to read and understand them. Thus, for their in-depth analysis, it is crucial to recover the damaged parts and properly translate them into modern languages.

To address these issues existing in historical records, 
we formulate them as the task of language modeling, especially for the recovery and neural machine translation, by leveraging the advanced neural networks. Moreover, we apply topic modeling to the translated historical records to efficiently discover the important historical events over the last hundreds of years.
In particular, we utilize two representative Korean historical records: the Annals of the Joseon Dynasty and the Diaries of the Royal Secretariat (hereafter we refer to them as \textbf{AJD} and \textbf{DRS} respectively).
These records, which contain 50 million and 243 million characters respectively, are recognized as the largest historical records in the world. Considering their high value, UNESCO recognized them as the Memory of the World.\footnote{\url{http://www.unesco.org/new/en/communication-and-information/memory-of-the-world/register/full-list-of-registered-heritage/registered-heritage-page-8/the-annals-of-the-choson-dynasty/}}\footnote{\url{http://www.unesco.org/new/en/communication-and-information/memory-of-the-world/register/full-list-of-registered-heritage/registered-heritage-page-8/seungjeongwon-ilgi-the-diaries-of-the-royal-secretariat/}} 
These two historical corpora contain the contents of five hundred years from the fourteenth century to the early twentieth century. In detail, AJD consists of administrative affairs with national events, and DRS contains events that occurred around the kings of the Joseon Dynasty. 
These corpora are valuable as they contain diverse information including international relations and natural disasters. 
In addition, the contents of the records are objective since the writing rules are strict that political intervention, even from the kings, is not allowed by their independent institution.

Although DRS contains a much larger amount of information than AJD, only 10--20\% of DRS has been translated into the modern Korean language by a few dozens of experts for the last twenty years. The complete translation of DRS is currently expected to additionally take more than 30--40 years if only human experts continue to translate them.
Applying the neural machine translation models into the historical records contains several issues.
First, the pre-trained models for Chinese are not suitable to DRS and AJD, mainly because of the differences between Hanja and the Chinese language.
In the past, Korean historiographers borrowed the Chinese character to write the sentences spoken by Koreans. As a result, diverse characters had been moderated or created, and considerable grammatical differences exist between the Chinese language and Hanja.
Furthermore, several parts of those records are damaged and require restoration as shown in Fig.~\ref{fig:damaged_documents}. Therefore, these damaged parts should be restored in order to translate them correctly.
In order to address these issues, we propose a model suitable for the historical documents using the self-attention mechanism.

Overall, we propose a novel multi-task approach to restore the damaged parts and translate the records into a modern language. Afterward, we extract the meaningful historical topics from the world’s largest historical records as shown in Fig.~\ref{fig:overview}. 
This study makes the following contributions:

\begin{itemize}[noitemsep,topsep=0pt,leftmargin=0.2in]
\item We design a model based on the self-attention mechanism with multi-task learning to restore and translate the historical records. Results demonstrate that our methods are effective in restoring the damaged characters and translating the records into a modern language.
\item We translate all the untranslated sentences in DRS. We believe that this dataset would be invaluable for researchers in various fields.\footnote{The codes, trained model, and datasets are accessible via \url{https://github.com/Kyeongpil/deep-joseon-record-analysis}.}
\item We present a case study that extracts meaningful historical events by applying topic modeling, highlighting the importance of analysis of historical documents.
\end{itemize}

\section{Related Work}
This work broadly incorporates three different tasks: document restoration, machine translation, and document analysis. Therefore, this section describes studies related to the restoration of damaged documents, neural machine translation, and the analysis of historical records.

\subsection{Neural Machine Translation}
Recently, neural machine translation (NMT) has achieved outstanding achievements. Based on the encoder-decoder architecture, the attention mechanism~\cite{bahdanau2015neural} significantly improves the performance of NMT, by calculating the target context vector in the current time step via dynamically combining the encoding vectors of source words. The self-attention-based networks~\cite{vaswani2017attention} consider the correlations among all word pairs in the source and target sentences. Based on the success of self-attention networks, Transformer architecture for language modeling has been proposed, showing the forefront performances~\cite{devlin2019bert,radford2019language}. Especially, the pre-training approaches further improve the performances, since they train the model robustly with several tasks using a large document corpus. In addition, lightweight models, such as ALBERT~\cite{lan2019albert}, are proposed to reduce the model size while preserving the model performance. However, as most of the recent approaches focus on pre-training with documents written in a modern language, the model for historical datasets does not exist. Therefore, we adopt a lightweight model in the same manner as ALBERT to efficiently reconstruct and translate millions of documents. 

Regarding the translation task for the historical documents, several studies attempt to translate the ancient Chinese documents into modern Chinese language~\cite{zhang2019automatic,liu2019ancient}. However, as they mainly attempt to translate
archaic characters into the modern language using paired corpus, they do not fully utilize the unpaired corpus. 
Therefore, we improve the performance of machine translation for historical corpora with multi-task learning with the translation and restoration tasks, which fully utilize the paired and unpaired corpora.

\begin{figure}[t]
\centering
\includegraphics[width=\columnwidth,trim=0 5 25 0,clip]{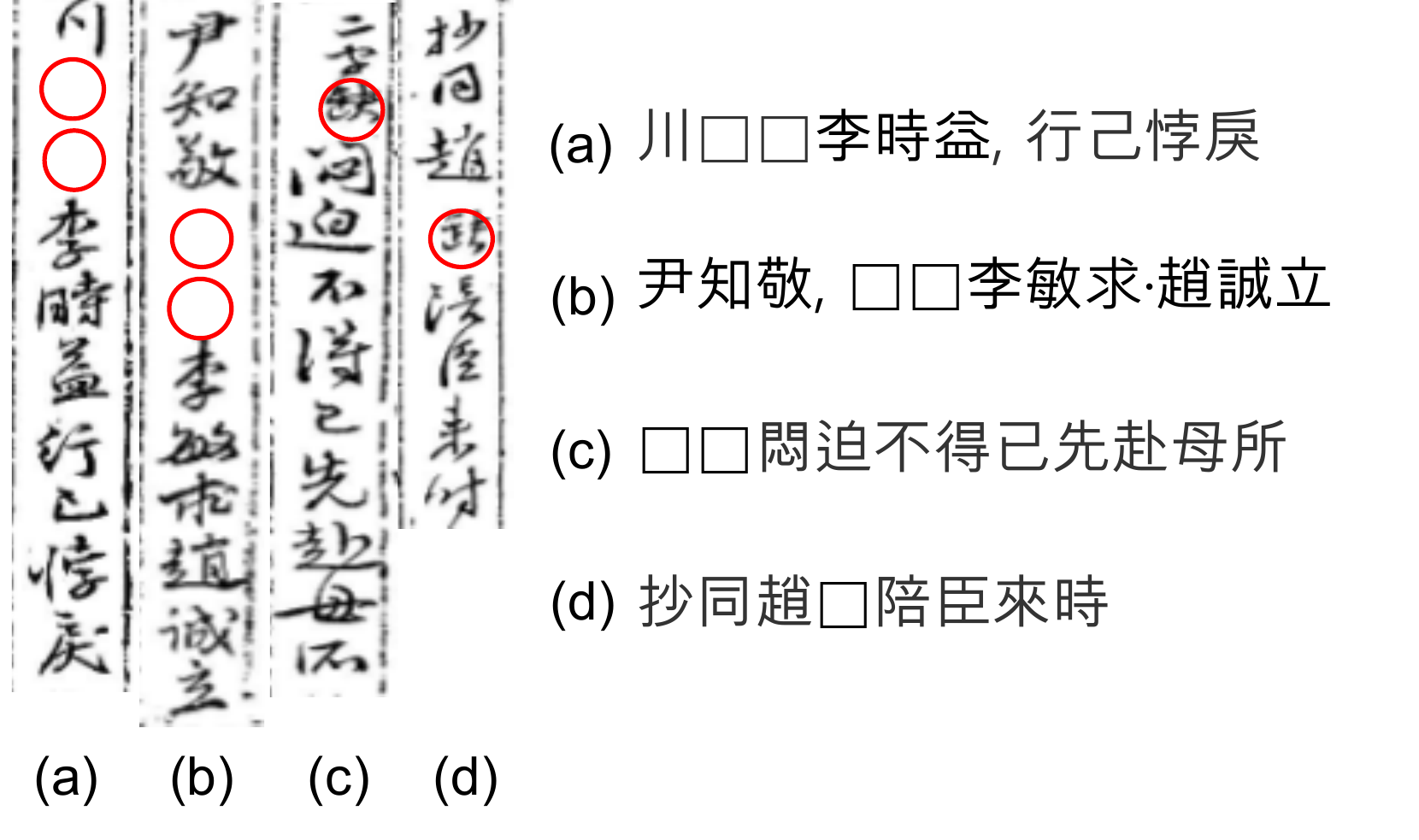}
\caption{Examples of damaged documents. Those characters that should be put in rectangles are damaged or unrecognizable.}
\label{fig:damaged_documents}
\end{figure}

\subsection{Restoration of Historical Documents}
Unfortunately, lots of characters in the historical records are damaged or misspelled. As shown in Fig.~\ref{fig:damaged_documents}, the damaged parts are prevalent in DRS, which significantly degrade the quality of subsequent translation tasks. To address this problem, several studies focus on normalizing the misspelled words~\cite{tang-etal-2018-evaluation,domingo2018spelling}, and others further apply language modeling to restore the parts of the documents via deep neural networks (DNNs)~\cite{caner2010shape,assael-etal-2019-restoring}.

Recently, the Cloze-style approach of machine reading comprehension (masked language modeling; MLM) predicts the original tokens for those positions where the words in the original sentence are randomly chosen and masked or replaced~\cite{hermann2015teaching}. Several studies significantly improved the model performance by pre-training the model via the Cloze-style approach. By utilizing the MLM approach with the self-attention mechanism and the large-scale training dataset, numerous models improve the performances of various downstream tasks including NMT task~\cite{baevski-etal-2019-cloze,devlin2019bert,zhang-etal-2019-ernie,conneau2019cross,liu2019roberta,clark2019electra}. However, to our knowledge, few studies apply such an MLM approach to restore the damaged parts.

Motivated by these studies, we design our model using masked language modeling based on the self-attention architecture to recover the damaged documents considering their contexts.

\subsection{Analysis on Historical Records}
Various studies apply the machine learning approaches to analyze the historical records~\cite{zhao2014sentiment,kumar2014ancient,mimno2012computational,kim2015simultaneous,bak2015five,bak2018conversational}. In addition, researchers adopt neural networks such as convolutional neural networks and autoencoders, for page segmentation and optical character recognition to convert the historical records in a digital form~\cite{chen2017convolutional,clanuwat2019kuronet}. Given such digital-form records, analysts attempt to utilize the topic modeling to discover the historically meaningful events~\cite{yang-etal-2011-topic}.

Especially, using the translated AJD, researchers discover historical events such as magnetic storm activities~\cite{yoo2015classification,hayakawa2017long}, meteors~\cite{lee2009orbital}, and solar activities~\cite{jeon2018relationship}. In political science, researchers analyze the decision patterns of a royal family in the Joseon Dynasty~\cite{bak2015five,bak2018conversational}. Besides, the dietary patterns and dynamic social relations among key figures during the Joseon Dynasty have been investigated~\cite{ki2018horse}. However, existing studies mainly rely on the documents translated by human experts. Therefore, we first translate the documents in AJD and DRS. Afterward, we apply topic modeling approaches to mine meaningful historical events over large-scale data.

\section{Proposed Methods}

\begin{figure}[t]
    \centering
    \includegraphics[width=0.84\columnwidth,trim=0 0 0 0,clip]{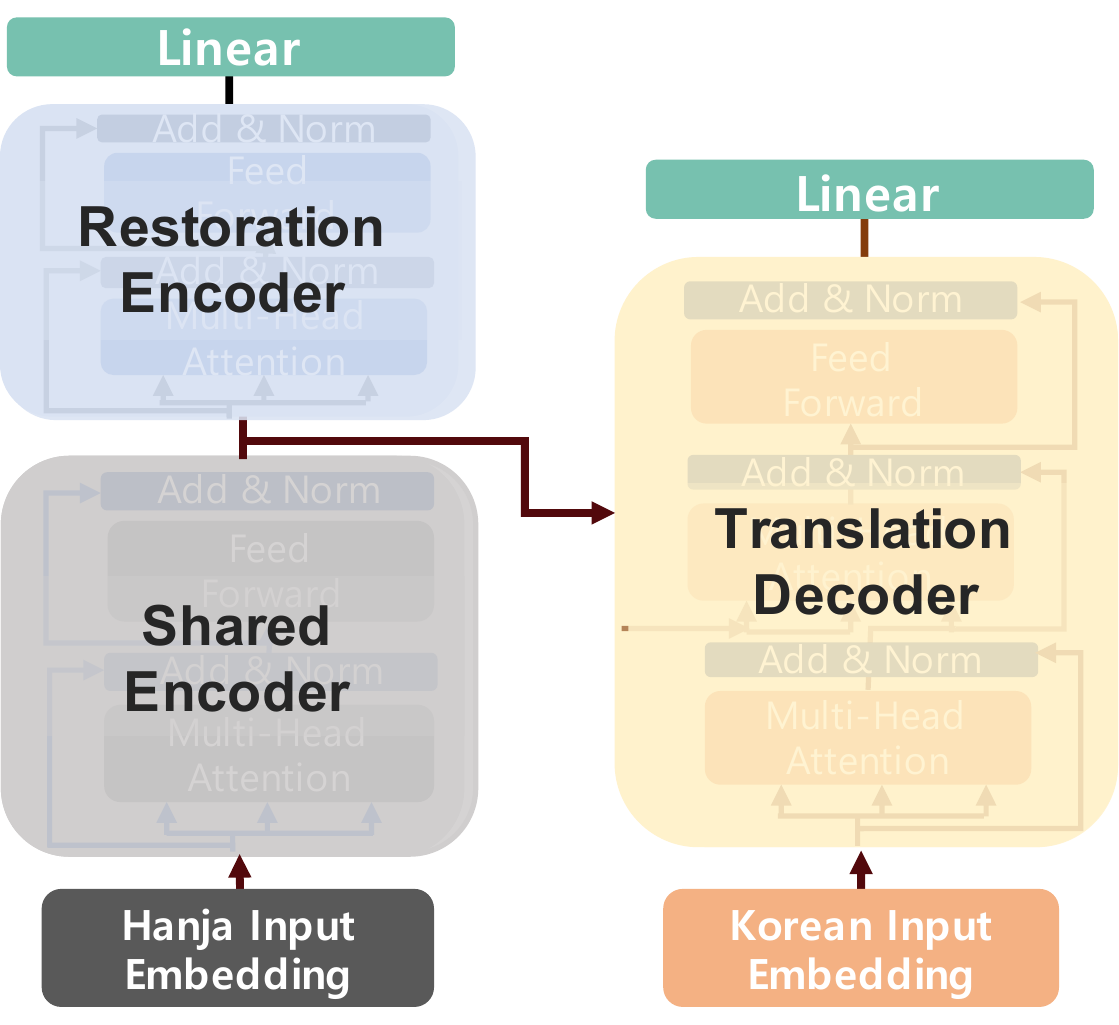}
    \caption{Overview of the proposed model for the restoration and translation tasks.}
    \label{fig:model}
    \vspace{-1em}
\end{figure}

This section describes a multi-task learning approach based on the Transformer networks to effectively restore and translate the historical records. The overview of our model is shown in Fig.~\ref{fig:model}.

AJD and DRS datasets consist of Hanja sentences $\mathcal{H}=\{h_1,\dots,h_N\}$ and Korean sentences $\mathcal{K}=\{k_1,\dots,k_N\}$, where each Korean sentence is translated from its corresponding Hanja sentence. Here, the Hanja represents the Chinese characters borrowed to write the Korean language in the past. Especially, DRS contains additional Hanja sentences $\widetilde{\mathcal{H}}=\{h_{N+1}, \dots,h_M\}$ that are not translated yet. Hence, we have in total $M$ Hanja sentences in the Hanja corpus such that $\hat{\mathcal{H}}=\mathcal{H}\cup\widetilde{\mathcal{H}}$ and $N$ Korean sentences in the Korean corpus $\mathcal{K}$.

Considering the properties of AJD and DRS, we design a multi-task learning approach with document restoration and machine translation, based on the Transformer networks.  As shown in Fig.~\ref{fig:model}, our model consists of embedding and output layers for Hanja and Korean, and three Transformer modules: the shared encoder, the restoration encoder, and the translation decoder. The restoration encoder is an encoder for the restoration task. The translation decoder is used for translating Hanja sentences into modern Korean sentences, and the shared encoder is used for both the restoration and translation tasks. By sharing the encoder module for both tasks, the shared encoder is trained with a large-scale corpus, i.e., the Hanja-Korean paired dataset and the additional unpaired Hanja dataset. The parameter sharing technique assists the model to learn rich information from the Hanja corpus. We apply the cross-layer parameter-sharing technique in the same manner as used in ALBERT~\cite{lan2019albert}, which shares the attention parameters for each Transformer encoder and decoder modules to reduce the model size and the inference time.

\subsection{Restoration of Damaged Documents}
The restoration task for damaged documents is similar to the MLM approach, which masks randomly chosen tokens in the input sentence and then predicts their original tokens in the corresponding position. We apply the MLM technique to restore the damaged documents, especially in the case of the Hanja sentences $\hat{\mathcal{H}}$.

For word indices $(w_1^{h_i}, \dots, w_{L_i}^{h_i})$ in the Hanja sentence $h_i$, where $L_i$ is the length of the $i$-th sequence, several words are randomly selected and replaced by a [MASK] token. We extract word embedding vectors $(e_1^{h_i}, \dots, e_{L_i}^{h_i}) \in \mathbb{R}^{d_{emb}}$ from the Hanja embedding layer combined with positional embedding vectors, where $d_{emb}$ represents the dimension size of the embedding space. Here, we apply the factorized embedding parameterization technique to reduce model parameters~\cite{lan2019albert}. These embedding vectors are projected onto the $d_{model}$-dimensional embedding space through a linear layer. Subsequently, the embedding vectors are transformed into the Hanja context vectors $(\hat{s}_1^{h_i}, \dots, \hat{s}_{L_i}^{h_i})$ via the shared encoder and the restoration encoder as
\begin{align}
\label{eq:shared_encoder}
s_1^{h_i}, \dots, s_{L_i}^{h_i} = f_S(e_1^{h_i}, \dots, e_{L_i}^{h_i}), \\
\hat{s}_1^{h_i}, \dots, \hat{s}_{L_i}^{h_i} = f_R(s_1^{h_i}, \dots, s_{L_i}^{h_i}),
\end{align}

where $f_S$ and $f_R$ functions represent the shared encoder and the restoration encoder, respectively. The Hanja context vectors is non-linearly transformed into the output vector $z_k^{h_i} \in \mathbb{R}^{d_{emb}}$ via the output layer. We also apply the factorized embedding parameterization technique to the output layers for parameter reduction. We calculate the probability $P(\hat{w}_{k,m}^{h_i} | w_1^{h_i}, \dots, w_{L_i}^{h_i})$ for the index $m$ of the original token $\hat{w}_k^{h_i}$, using the softmax function as
\begin{equation}
\small
P(\hat{w}_{k,m}^{h_i}|w_1^{h_i}, \dots, w_{L_i}^{h_i}) = \frac{\exp({\mathbf{W}_m^h}^{\top} z_k^{h_i})}{\sum_{j}^{|V_h|}{\exp({\mathbf{W}_j^h}^{\top} z_k^{h_i})}},
\end{equation}
where $|V_h|$ is the size of the Hanja vocabulary.

\subsection{Neural Machine Translation for Historical Records}
In order to facilitate the training of our translation module, we exploit the Hanja-Korean paired dataset $\{(h_i, k_i) | h_i\in\mathcal{H}, k_i\in\mathcal{K} \}$. As shown in Fig.~\ref{fig:model}, we first extract the Hanja context vectors $(s_1^{h_i},\dots, s_{L_i}^{h_i})$ from the word tokens in the Hanja sentence $h_i$, using the shared encoder in the same manner as in Eq.~\ref{eq:shared_encoder}. Utilizing the Hanja context vectors and previously predicted Korean words $(w_1^{k_i}, \dots, w_{t-1}^{k_i})$, we subsequently calculate the $d_{model}$-dimensional Korean context vector $s_t^{k_i}$ for the current time step $t$ as
\begin{equation}
s_t^{k_i}=f_D(s_1^{h_i},\dots,s_{L_i}^{h_i},w_1^{k_i},\dots,w_{t-1}^{k_i}),
\end{equation}
where $f_D$ represents the translation decoder layers. After calculating the Korean context vector $s_t^{k_i}$, we non-linearly transform the context vector to the output vector $z_t^{k_i} \in \mathbb{R}^{d_{emb}}$, through the output layer, along with the above-mentioned factorized embedding parameterization for parameter reduction. Finally, we yield the probability that the word $V_m$ is generated from the $t$-th step as
\begin{equation}
\small
P(w_{t,m}^{k_i}|h_i,w_{1:t-1}^{k_i}) = \frac{\exp({\mathbf{W}_m^k}^{\top} z_t^{k_i})}{\sum_{j}^{|V_k|}{\exp({\mathbf{W}_j^k}^{\top} z_t^{k_i})}},
\end{equation}
where $|V_k|$ is the size of the vocabulary for the Korean corpus, and $\mathbf{W}^k\in\mathbb{R}^{|V_k|\times d_{emb}}$ is the output layer for the Korean corpus.

As previously mentioned, we employ the parameter sharing approach for the encoder module, (i.e., the shared encoder), thus enhancing the robustness of our model, especially with the Hanja dataset.

\subsection{Training and Inference}
In order to train our model, we use the cross-entropy loss to maximize the probability of the original token indices for the masked tokens and the target sentence for the translation task as
\begin{align}
\resizebox{0.85\hsize}{!}{$\mathcal{L}_{rst} = -\frac{1}{M}\sum_{h_i \in \hat{\mathcal{H}}}{\mathbb{E}_{k \sim \xi(h_i)}\big[\log P(w_k^{h_i}|h_i)\big]}$}, \\
\resizebox{0.87\hsize}{!}{
$\mathcal{L}_{trs}=-\frac{1}{N}\sum_{i=1}^{N}{\big[ \frac{1}{|k_i|}\sum_{t=1}^{|k_i|}{P(w_t^{k_i}|h_i,w_{1:t-1}^{k_i})}\big]}$},
\end{align}
where $\xi(\cdot)$ is an operator that randomly selects the tokens from each sentence for MLM. In this study, we apply not only unigram masking but also the n-gram masking techniques (i.e., bigrams and trigrams), as previously applied~\cite{zhang-etal-2019-ernie}. Finally, the total loss is defined as 
\begin{align}
\small
\mathcal{L} = \mathcal{L}_{rst} + \mathcal{L}_{trs}.
\end{align}

Our model is optimized by using the rectified Adam~\cite{liu2019variance} with the layer-wise adaptive rate scheduling technique~\cite{you2017large}. We also apply the gradient accumulation technique and update our model for each loss asynchronously, to increase the batch size and efficiently manage the GPU memory.

After training the model, the damaged tokens are replaced by the [MASK] token during the restoration stage, and the model obtains the top-K characters with the highest probabilities, among which users can choose and confirm a correct characters in the position of the damaged parts. In addition, we translate all the Hanja records that are not yet translated for further in-depth analysis. When translating the Hanja sentence, we additionally apply beam search with length normalization. The translation task for all the untranslated records using 20 V100 GPUs had a duration of approximately five days.

\section{Experiments}
This section first describes our datasets and experimental settings.

\subsection{Datasets and Preprocessing}

\begin{table}[t]
\centering
\small
\setlength{\tabcolsep}{4.7pt}
\begin{tabular}{c|c|c|c}
\hline
& Paired Hanja & Unpaired Hanja & Korean \\
\hline
\#(Train data) & 239,226 & 1,377,320 & 239,226 \\
\#(Test data) & 20,000 & 20,000 & 20,000 \\
1st Quartile & 26 & 27 & 22 \\
Mean & 143.81 & 165.66 & 123.68 \\
3rd Qquatile & 106 & 113 & 80 \\
Median & 52 & 55 & 40 \\
Vocab size & 8,742 & 8,742  & 24,000 \\
\hline
\end{tabular}
\caption{Dataset summary. The third to the sixth rows indicate the statistics for the length of each document.}
\vspace{-1em}
\label{tab:dataset}
\end{table}

To train our model, we collect most of the documents of AJD and DRS, including those manually translated to date, provided by the National Institute of the Korean History\footnote{\url{http://www.history.go.kr/}}. The records contain approximately 250K documents for AJD and 1.4M documents for DRS.

After collecting documents, we tokenize each Hanja sentence into the character-level tokens, similar to previous studies~\cite{zhang2014character,li-etal-2018-simple}, and also tokenize each Korean sentence based on the unigram language model~\cite{kudo-2018-subword} provided by Google's SentencePiece library.\footnote{\url{https://github.com/google/sentencepiece}} Here, we included those words appearing more than ten times in the Hanja vocabulary, the size of which is about 8.7K words. For the Korean corpus, we limit the size of the Korean vocabulary to 24K. The out-of-vocabulary words are replaced with UNK (unknown) tokens. To improve the stability and efficiency during the training stage, we filter out those Hanja sentences with less than four tokens or more than 350 tokens and those Korean sentences with less than four tokens or more than 300 tokens. Note that the portion of sentences filtered out from each dataset is less than 10\%.

To evaluate the performance of our model, we randomly select 20K sentences as a test dataset for each of the paired and the unpaired sets. The sizes of the training set for the Hanja-Korean paired corpus and the unpaired Hanja corpus are 240K and 1.38M, respectively. The statistics of the dataset are summarized in Table~\ref{tab:dataset}.

\subsection{Hyper-parameter Settings}
We set hyper-parameters similarly to the BERT~\cite{devlin2019bert} base model. We set the size of the embedding dimension $d_{emb}$, the hidden vector dimension $d_{model}$, and the dimension of the position-wise feed-forward layers as 256, 768, and 3,072, respectively. The shared encoder, the translation decoder, and the restoration encoder consist of 12, 12, and 6 layers, respectively. We use 12 attention heads for each multi-head attention layer. Overall, the total number of parameters is around 168.8M.

\subsection{Mining Historical Records via Topic Modeling}
\label{sec:experience_topic_modeling}
After obtaining machine-translated outputs of the remaining records, we apply topic modeling to the full set of documents for exploratory analysis of historical events. To be specific, the full set of documents include all of the manually translated records as well as machine-translated records by our model. By using each translated record $k_i$ and its written date information $d_i$, we first parse the document into morphemes and then use the only noun and adjective tokens. Afterward, we build the term-date matrix $\mathbf{V} \in \mathbb{R}^{V \times D}$ where $V$ is the vocabulary size and $D$ is the number of dates in the total set of historical documents.

In this study, we utilize non-negative matrix factorization (NMF)~\cite{lee2001algorithms} as a topic modeling method\footnote{Topic modeling includes several methods such as latent Dirichlet allocation (LDA)~\cite{blei2003latent}-based and nonnegative matrix factorization-based models~\cite{lee2001algorithms}. We additionally tested topic modeling with LDA, but the results of NMF are slightly better than those of LDA.}. We first assume that there exist $K$ topics in the corpus. The term-date matrix $\mathbf{V}$ is decomposed into the term-topic weight matrix $\mathbf{W} \in \mathbb{R}^{V \times K}$ and the date-topic weight matrix $\mathbf{H} \in \mathbb{R}^{D \times K}$ as
\begin{equation}
\label{eq:nmf}
\small
    \mathbf{W}, \mathbf{H} = \argmin_{\mathbf{W}, \mathbf{H} \geq 0}{{\lVert \mathbf{V} - 
    \mathbf{WH}^{\top}\rVert}_F^2 + \alpha \cdot \psi( \mathbf{W}, \mathbf{H})},
\end{equation}
where ${\lVert \cdot \rVert}_F$ represents the Frobenius norm, and $\psi$ and $\alpha$ represent the $L_1$ regularization function and the regularization weight, respectively. We set the number of topics $K$ as 20\footnote{We set the number of topics as 20 after we conducted experiments by varying the topic numbers, such as 10, 15, 20, 30, and 50.} and the regularization weight $\alpha$ as 0.1.

\section{Experimental Results}
This section describes the results of the performances of our model for restoration and translation, followed by qualitative examples of each task as well as topic modeling results.

\subsection{Document Restoration}

\begin{CJK*}{UTF8}{bsmi}
\begin{table*}[t]
    \centering
    \begin{tabular}{c|c}
        \hline
        Original &  上在慶德\textbf{\textcolor{blue}{宮}}. 停常\textbf{\textcolor{blue}{參}}·經筵. \\
        Predicted & 上在慶熙\textbf{\textcolor{red}{宮}}. 停常\textbf{\textcolor{red}{參}}·經筵. \\
        \hline
        Original &  右承旨李世用疏曰云云. 省\textbf{\textcolor{blue}{疏具}}悉. 疏辭, 下該曹稟處. \\
        Predicted & 右承旨李世用疏曰云云. 省\textbf{\textcolor{red}{疏具}}悉. 疏辭, 下該曹稟處. \\
        \hline
        Original &  \textbf{\textcolor{blue}{玉堂箚}}子. 答曰, 省箚具悉. 箚辭當採用焉. 內下記草 \\
        Predicted & \textbf{\textcolor{red}{玉堂箚}}子. 答曰, 省箚具悉. 箚辭當採用焉. 內下記草 \\
        \hline
        Original &  又啓曰, 假注書\textbf{\textcolor{blue}{金基龍}}, 身病猝重, 勢難察任, 今姑改差, 何如? 傳曰, 允. \\
        Predicted & 又啓曰, 假注書\textbf{\textcolor{red}{李基淳}}, 身病猝重, 勢難察任, 今姑改差, 何如? 傳曰, 允. \\
        \hline
    \end{tabular}
    \vspace{-0.5em}
    \caption{Our model prediction results. Blue- and red-colored letters represent masked and predicted ones, respectively.}
    \vspace{-0.5em}
    \label{tab:restoration_examples}
\end{table*}
\end{CJK*}

\begin{table}[t]
    \centering
    \setlength{\tabcolsep}{4.7pt}
    \begin{tabular}{c|c|c|c}
         \hline
         &  HITS@1 & HITS@5 & HITS@10 \\
         \hline
         Baseline & 77.83\% & 88.29\% & 90.89\% \\
         Full model & 75.20\% & 86.21\% & 89.09\% \\
         \hline
    \end{tabular}
    \caption{Top-K accuracies for the restoration task.}
    \vspace{-1em}
    \label{tab:restoration_performance}
\end{table}

We evaluate the performance of our model on the document restoration task on the test dataset. We also compare performance between the model trained with and without multi-task learning. Table~\ref{tab:restoration_performance} shows the results of top-K (HITS@K). The top-10 accuracy of our proposed model is almost 89\%, which indicates the high performance of our model and demonstrates that our model provides analysts with appropriate options. However, the baseline model, trained without multi-task learning, performs slightly better than the one with multi-task learning. This shows that the baseline model is more specialized in the document restoration task. However, although our model performance is slightly lower than the baseline model, the benefits of the multi-task learning approach are significantly manifested in the NMT task as shown in Table~\ref{tab:translation_performance}. As our model shows the acceptable performances on both the restoration and the translation tasks, we conclude that our model learns the purpose of our research well via multi-task learning. We will further discuss the main benefits of multi-task learning in Section~\ref{sec:resullts_mt}.

We further investigate the qualitative results of the document restoration task. Table~\ref{tab:restoration_examples} shows four randomly sampled, example pairs. As shown in the first three rows of this table, the model also has the ability to predict bi-gram and tri-gram character-level tokens because the model is trained using n-gram-based MLM. Furthermore, although each character is not exactly the same as the original one, the last example in the table shows that our model restores the proper format of the name part. However, predicting the exact name is a difficult task for human experts, even when considering the context of the sentence, as prior knowledge is necessary to predict the exact name.
Therefore, we quantitatively measured the model performance on the proper nouns, e.g. person and location names, using 200 samples of them. The average top-10 accuracy is only 8.3\%, significantly lower than the overall accuracy, which is larger than 89\%. We conjecture that the degradation is mainly due to the difficulty in maintaining the information of the proper nouns, which would require external knowledge. We leave it as our future work.

\subsection{Machine Translation Quality}
\label{sec:resullts_mt}

\begin{table*}[t]
    \centering
    \begin{tabularx}{\textwidth}{l|X}
        \hline
        Original & 上在昌慶宮. 停常參·經筵. \\
        Predicted & 상이 창경궁에 있었다. 상참과 경연을 정지하였다. \\
        Predicted (Eng.) & King was in the Changkyeong palace. He stopped the discussion of political affairs with other officers. \\
        \hline
        Original & 答大司憲南龍翼疏曰, 省疏具悉. 內局提調之任, 當勉副, 卿其勿辭, 救護母病, 從速上來察職. \\
        Predicted & 대사헌 남용익의 상소에 답하기를, \enquote{상소를 보고 잘 알았다. 내국 제조의 직임은 부지런히 마지못해 경의 뜻을 따라주니, 경은 사직하지 말고 어미를 구호하는 데에 속히 올라와 직임을 살피라.} 하였다. \\
        Predicted (Eng.) & Replying to the Prosecutor General Namyongik's memorial, the king said, \enquote{I looked at the memorial and thoroughly understood what it meant. As the position of the director at the office of the royal physicians cannot help but agree to your message, you should not resign your position, care for your mother's illness, and come back to be responsible for your duties quickly.} \\
        \hline
        Original &  夜一更, 月暈. 五更, 西方坤方, 有氣如火光. \\
        Predicted & 밤 1경에 달무리가 졌다. 5경에 서방, 곤방에 화광 같은 기운이 있었다. \\
        Predicted (Eng.) & The moon has a ring around it at 7-9 PM. At 3-5 AM, there was the light of the fire in the west and south-west. \\
        \hline
    \end{tabularx}
    \caption{Examples of original Hanja sentences, ground-truth sentences, and predicted sentences. For readability, we appended English sentences corresponding to the predicted sentences in each row.}
    \label{tab:translation_examples}
\end{table*}

\begin{table}[t]
    \centering
    \begin{tabular}{c|c|c|c}
         \hline
         &  BLEU & METEOR &  ROUGE-L \\
         \hline
         Base (1) & 0.3547 & 0.3488 & 0.6082 \\
         Base (3) & 0.3536 & 0.3482 & 0.6127 \\
         \hline
         Full (1) & 0.5269 & 0.4594 & 0.7463 \\
         Full (3) & \textbf{0.5410} & \textbf{0.4719} & \textbf{0.7606} \\
         \hline
    \end{tabular}
    \caption{Results of the performance of the translation task. \enquote{Base} and \enquote{Full} represent the model trained only using the machine translation task and the model trained using multi-task learning with machine translation and restoration tasks, respectively.}
    \label{tab:translation_performance}
\end{table}

To investigate the performance of the machine translation task, we translate the Hanja sentences in the test dataset and then evaluate the model performance. As shown in Table~\ref{tab:translation_performance}, the results for the translation task are evaluated by BLEU~\cite{papineni2002bleu}, METEOR~\cite{banerjee2005meteor}, and ROUGE-L~\cite{lin-2004-rouge}. In this result, \enquote{Full} represents our proposed model trained by multi-task learning of the translation and the restoration tasks. Therefore, the model is trained to take both the translated and untranslated sentences. On the other hand, \enquote{Base} represents the model trained only by the translation task, and thus, the model is trained to accept only the translated sentences. Our model outperforms the baseline model with a significant margin. 

Furthermore, we generate sentences using the beam search method with the length normalization. In this study, we compare the greedy search and the beam search with a beam size of 3. As shown in Table~\ref{tab:translation_performance}, results obtained with a beam size of 3 are slightly better than the greedy search method. Finally, the BLEU score of our model is obtained as 0.5410, which indicates that our model performs reasonably well, compared to other recent models trained in other languages.

\begin{table}[t]
    \centering
    \begin{tabular}{c|c|c|c}
         \hline
         &  Multi-task & Scratch & Pipelining \\
         \hline
         BLEU & \textbf{0.5410} & 0.3536 & 0.3755 \\
         \hline
    \end{tabular}
    \caption{BLEU scores of the models trained with multi-task learning, scratch, and the pretraining-then-finetuning approach (pipelining), respectively.}
    \vspace{-1em}
    \label{tab:translation_pipelining}
\end{table}

We additionally compared our model to the model trained via the pretraining-then-finetuning approach. As shown in Table~\ref{tab:translation_pipelining}, the BLEU score of this approach is 0.3755, which is 5.9\% higher than that of the model trained from scratch but 28.7\% lower than our multi-task learning approach. The results can be explained for two reasons. First, as the size of unpaired data is much larger than that of paired data, the multi-task learning fully utilizes the paired and unpaired data for the translation task, compared to the pretraining-then-finetuning approach. Second, The pretraining-then-finetuning approach has a catastrophic forgetting problem~\cite{chen-etal-2020-recall}. In other words, the finetuning step can fail to maintain the knowledge acquired at the pretraining step. However, as both reconstruction and translation tasks are crucial for historical documents, such a forgetting issue is critical to our tasks.

We also tested the quality of the Hanja-Korean translation task using a Chinese-Korean machine translator. As few publicly available machine translation models for Chinese-Korean exist, we used Google Translate\footnote{\url{https://translate.google.co.kr}} instead. The translator failed to translate given Hanja sentences in most cases, mainly because Hanja and Chinese have different properties in terms of grammar and word meanings.

To investigate the translation performance qualitatively, we sampled translated samples. Table~\ref{tab:translation_examples} shows the sentences translated from the untranslated documents by our model. For readability, we append English sentences corresponding to the predicted sentences in each row. Each result indicates that our model generates the modern sentences corresponding to contexts of the source Hanja sentences. Interestingly, the third example in the table is related to the astronomical observation of the aurora. Later, we found prior studies confirming that the red energy mentioned in our document was an aurora~\cite{zhang1985korean,stephenson2008vapours}. This highlights the importance of the machine translation task of the historical records, as it is essential to survey by researchers in various fields such as astrophysics and geology. Therefore, we further analyze the documents with the topic modeling approach.

\subsection{Results of Topic Modeling}

\begin{figure*}[t]
    \centering
    \includegraphics[width=0.96\textwidth,trim=0 5 0 0,clip]{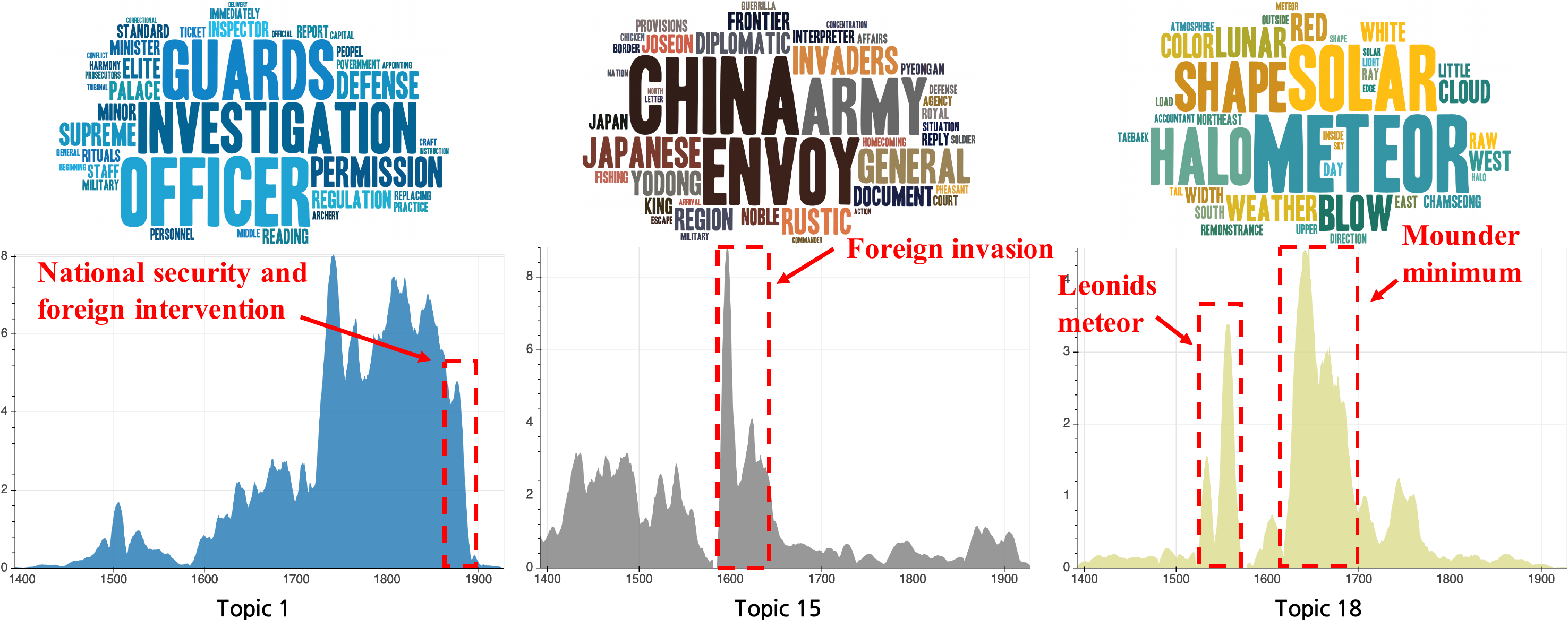}
    \caption{Three topics extracted from topic modeling. We translated topic keywords into English for readability.}
    \label{fig:topic_result}
\end{figure*}

As described in Section~\ref{sec:experience_topic_modeling}, we calculate the term-topic weight matrix $\mathbf{W}$ and date-topic weight matrix $\mathbf{H}$. We select three interesting topics from the total of $K$ topics and visualize the term-topic weights in $\mathbf{W}$ using the word cloud and the date-topic matrix $\mathbf{H}$ in a smoothed time-series graph for each topic. Fig.~\ref{fig:topic_result} shows the results.

The first topic is related to troops and military exercise. As shown in the red dashed box in the time-series graph, the weights dramatically decrease in 1882, while the weights continuously increase after the biggest war in 1592. In fact, a coup attempt of the old-fashioned soldiers occurred in 1882, causing the national intervention of neighboring countries and the decline of self-reliant defense. The fifteenth topic is related to war and national defense. Although this topic is related to the preceding military topic, it is more related to the international relationship compared to the first one. In the early years of the dynasty, northern enemies and pirates frequently invaded Joseon, which reveals as the large topical weights in the beginning. The weights increase in the late sixteenth century, and the weight maintains at a high level until 1637 when three great wars broke out in Joseon.

The eighteenth topic is related to astronomical observations such as a halo and a meteor shower. In the mid-sixteenth century, people observed the Leonids, as shown in the first red box of the graph. We later found that experts in astronomy also discovered this in the past, using AJD~\cite{yang2005analysis}. Moreover, from the mid-seventeenth century to the early eighteenth century, the number of sunspots was low. Solar observers name this event as the Maunder minimum~\cite{eddy1976maunder,shindell2001solar}. This event caused abnormal climate phenomena, such as the third example in Table~\ref{tab:translation_examples}, as shown in the second red box of the graph. This topic demonstrates the importance of the use of historical records since it is difficult to easily spot the phenomena that occurred centuries ago.

Note that previous studies mainly attempted to exploit only AJD or translated parts of DRS. However, we utilize both AJD and the majority of DRS records by applying advanced NMT techniques. 
When performing topic modeling by using only those manually translated sentences, it failed to include topics such as the health of the royal families and actions against treason sinners, which were revealed by our approach. It is because the voluminous documents that have not been manually translated contain their own topics.
Thereby, we extract several valuable topics even with no special knowledge in the Hanja domain. Translating the historical records into modern languages expands our knowledge base, and analysis of the records using machine translation and text mining techniques may help the analysts effectively explore the historical records.

\section{Conclusions}
In this paper, we proposed a novel approach to translate and restore the historical records of the Joseon dynasty by formulating the multi-task learning task based on the self-attention mechanism. Our approach significantly increases the translation quality by learning the rich contents in large documents. 
We anticipate these tasks are the first steps towards translating the ancient Korean historical records into modern languages such as English. Furthermore, the model effectively predicts the original words from the damaged parts of the documents, which is an essential step for restoaring the damaged documents. 
Results from text mining approaches show that our approaches have the potential in supporting analysts in effectively exploring the large volume of historical documents.
We also expect researchers from diverse domains can explore documents and discover historical findings such as astronomical phenomena and undiscovered international affairs, with no special domain knowledge. 
As future work, we will also leverage the transfer learning approach to translate historical documents into other languages, such as English or French. We also plan to apply knowledge graph-based machine learning approaches, e.g. knowledge graph embedding and graph neural networks, to discover historical events and relations.

\section*{Acknowledgements}
This work was supported by Institute for Information \& communications Technology Planning \& Evaluation (IITP) grant funded by the Korea government (MSIT) (No.2020-0-00368, A Neural-Symbolic Model for Knowledge Acquisition and Inference Techniques, No.2019-0-00075, Artificial Intelligence Graduate School Program (KAIST), and No.2021-0-01341, Artificial Intelligence Graduate School Program (Chung-Ang University)) and the National Research Foundation of Korea (NRF) grant funded by the Korean government (MSIT) (No.NRF-2019R1A2C4070420).


\bibliography{ref_abbr,reference,anthology}

\begin{thebibliography}{47}
\expandafter\ifx\csname natexlab\endcsname\relax\def\natexlab#1{#1}\fi

\bibitem[{Assael et~al.(2019)Assael, Sommerschield, and
  Prag}]{assael-etal-2019-restoring}
Yannis Assael, Thea Sommerschield, and Jonathan Prag. 2019.
\newblock \href {https://doi.org/10.18653/v1/D19-1668} {Restoring ancient text
  using deep learning: a case study on {G}reek epigraphy}.
\newblock In \emph{Proceedings of the 2019 Conference on Empirical Methods in
  Natural Language Processing and the 9th International Joint Conference on
  Natural Language Processing (EMNLP-IJCNLP)}, pages 6368--6375, Hong Kong,
  China. Association for Computational Linguistics.

\bibitem[{Baevski et~al.(2019)Baevski, Edunov, Liu, Zettlemoyer, and
  Auli}]{baevski-etal-2019-cloze}
Alexei Baevski, Sergey Edunov, Yinhan Liu, Luke Zettlemoyer, and Michael Auli.
  2019.
\newblock \href {https://doi.org/10.18653/v1/D19-1539} {Cloze-driven
  pretraining of self-attention networks}.
\newblock In \emph{Proceedings of the 2019 Conference on Empirical Methods in
  Natural Language Processing and the 9th International Joint Conference on
  Natural Language Processing (EMNLP-IJCNLP)}, pages 5360--5369, Hong Kong,
  China. Association for Computational Linguistics.

\bibitem[{Bahdanau et~al.(2015)Bahdanau, Cho, and Bengio}]{bahdanau2015neural}
Dzmitry Bahdanau, Kyunghyun Cho, and Yoshua Bengio. 2015.
\newblock Neural machine translation by jointly learning to align and
  translate.
\newblock In \emph{Proc. the International Conference on Learning
  Representations (ICLR)}.

\bibitem[{Bak and Oh(2015)}]{bak2015five}
JinYeong Bak and Alice Oh. 2015.
\newblock Five centuries of monarchy in korea: mining the text of the annals of
  the joseon dynasty.
\newblock In \emph{Proceedings of the SIGHUM Workshop on Language Technology
  for Cultural Heritage, Social Sciences, and Humanities}.

\bibitem[{Bak and Oh(2018)}]{bak2018conversational}
JinYeong Bak and Alice Oh. 2018.
\newblock Conversational decision-making model for predicting the king’s
  decision in the annals of the joseon dynasty.
\newblock In \emph{Proc. of the Conference on Empirical Methods in Natural
  Language Processing (EMNLP)}.

\bibitem[{Banerjee and Lavie(2005)}]{banerjee2005meteor}
Satanjeev Banerjee and Alon Lavie. 2005.
\newblock Meteor: An automatic metric for mt evaluation with improved
  correlation with human judgments.
\newblock In \emph{Proceedings of the Proc. the Annual Meeting of the
  Association for Computational Linguistics (ACL) workshop on intrinsic and
  extrinsic evaluation measures for machine translation and/or summarization}.

\bibitem[{Blei et~al.(2003)Blei, Ng, and Jordan}]{blei2003latent}
David~M Blei, Andrew~Y Ng, and Michael~I Jordan. 2003.
\newblock Latent dirichlet allocation.
\newblock \emph{Journal of machine Learning research}.

\bibitem[{Caner and Haritaoglu(2010)}]{caner2010shape}
Gulcin Caner and Ismail Haritaoglu. 2010.
\newblock Shape-dna: effective character restoration and enhancement for arabic
  text documents.
\newblock In \emph{International Conference on Pattern Recognition}.

\bibitem[{Chen et~al.(2017)Chen, Seuret, Hennebert, and
  Ingold}]{chen2017convolutional}
Kai Chen, Mathias Seuret, Jean Hennebert, and Rolf Ingold. 2017.
\newblock Convolutional neural networks for page segmentation of historical
  document images.
\newblock In \emph{International Conference on Document Analysis and
  Recognition}.

\bibitem[{Chen et~al.(2020)Chen, Hou, Cui, Che, Liu, and
  Yu}]{chen-etal-2020-recall}
Sanyuan Chen, Yutai Hou, Yiming Cui, Wanxiang Che, Ting Liu, and Xiangzhan Yu.
  2020.
\newblock \href {https://doi.org/10.18653/v1/2020.emnlp-main.634} {Recall and
  learn: Fine-tuning deep pretrained language models with less forgetting}.
\newblock In \emph{Proceedings of the 2020 Conference on Empirical Methods in
  Natural Language Processing (EMNLP)}, pages 7870--7881, Online. Association
  for Computational Linguistics.

\bibitem[{Clanuwat et~al.(2019)Clanuwat, Lamb, and
  Kitamoto}]{clanuwat2019kuronet}
Tarin Clanuwat, Alex Lamb, and Asanobu Kitamoto. 2019.
\newblock Kuronet: Pre-modern japanese kuzushiji character recognition with
  deep learning.
\newblock In \emph{2019 International Conference on Document Analysis and
  Recognition (ICDAR)}.

\bibitem[{Clark et~al.(2019)Clark, Luong, Le, and Manning}]{clark2019electra}
Kevin Clark, Minh-Thang Luong, Quoc~V Le, and Christopher~D Manning. 2019.
\newblock Electra: Pre-training text encoders as discriminators rather than
  generators.
\newblock In \emph{Proc. the International Conference on Learning
  Representations (ICLR)}.

\bibitem[{Conneau and Lample(2019)}]{conneau2019cross}
Alexis Conneau and Guillaume Lample. 2019.
\newblock Cross-lingual language model pretraining.
\newblock In \emph{Proc. the Advances in Neural Information Processing Systems
  (NIPS)}.

\bibitem[{Devlin et~al.(2019)Devlin, Chang, Lee, and
  Toutanova}]{devlin2019bert}
Jacob Devlin, Ming-Wei Chang, Kenton Lee, and Kristina Toutanova. 2019.
\newblock Bert: Pre-training of deep bidirectional transformers for language
  understanding.
\newblock In \emph{NAProc. the Annual Meeting of the Association for
  Computational Linguistics (ACL)}.

\bibitem[{Domingo and Nolla(2018)}]{domingo2018spelling}
Miguel Domingo and Francisco~Casacuberta Nolla. 2018.
\newblock Spelling normalization of historical documents by using a machine
  translation approach.
\newblock In \emph{Proceedings of the Conference of the European Association
  for Machine Translation}.

\bibitem[{Eddy(1976)}]{eddy1976maunder}
John~A Eddy. 1976.
\newblock The maunder minimum.
\newblock \emph{Science}.

\bibitem[{Hayakawa et~al.(2017)Hayakawa, Iwahashi, Ebihara, Tamazawa, Shibata,
  Knipp, Kawamura, Hattori, Mase, Nakanishi et~al.}]{hayakawa2017long}
Hisashi Hayakawa, Kiyomi Iwahashi, Yusuke Ebihara, Harufumi Tamazawa, Kazunari
  Shibata, Delores~J Knipp, Akito~D Kawamura, Kentaro Hattori, Kumiko Mase,
  Ichiro Nakanishi, et~al. 2017.
\newblock Long-lasting extreme magnetic storm activities in 1770 found in
  historical documents.
\newblock \emph{The Astrophysical Journal}.

\bibitem[{Hermann et~al.(2015)Hermann, Ko{\v{c}}isk{\`y}, Grefenstette,
  Espeholt, Kay, Suleyman, and Blunsom}]{hermann2015teaching}
Karl~Moritz Hermann, Tom{\'a}{\v{s}} Ko{\v{c}}isk{\`y}, Edward Grefenstette,
  Lasse Espeholt, Will Kay, Mustafa Suleyman, and Phil Blunsom. 2015.
\newblock Teaching machines to read and comprehend.
\newblock In \emph{Proc. the Advances in Neural Information Processing Systems
  (NIPS)}.

\bibitem[{Jeon et~al.(2018)Jeon, Noh, and Lee}]{jeon2018relationship}
Junhyeok Jeon, Sung-Jun Noh, and Dong-Hee Lee. 2018.
\newblock Relationship between lightning and solar activity for recorded
  between ce 1392--1877 in korea.
\newblock \emph{Journal of Atmospheric and Solar-Terrestrial Physics}.

\bibitem[{Ki et~al.(2018)Ki, Shin, Woo, Lee, Hong, and Shin}]{ki2018horse}
Ho~Chul Ki, Eun-Kyoung Shin, Eun~Jin Woo, Eunju Lee, Jong~Ha Hong, and
  Dong~Hoon Shin. 2018.
\newblock Horse-riding accidents and injuries in historical records of joseon
  dynasty, korea.
\newblock \emph{International journal of paleopathology}.

\bibitem[{Kim et~al.(2015)Kim, Choo, Kim, Reddy, and
  Park}]{kim2015simultaneous}
Hannah Kim, Jaegul Choo, Jingu Kim, Chandan~K Reddy, and Haesun Park. 2015.
\newblock Simultaneous discovery of common and discriminative topics via joint
  nonnegative matrix factorization.
\newblock In \emph{Proc. the ACM SIGKDD International Conference on Knowledge
  Discovery and Data Mining (KDD)}.

\bibitem[{Kudo(2018)}]{kudo-2018-subword}
Taku Kudo. 2018.
\newblock \href {https://doi.org/10.18653/v1/P18-1007} {Subword regularization:
  Improving neural network translation models with multiple subword
  candidates}.
\newblock In \emph{Proceedings of the 56th Annual Meeting of the Association
  for Computational Linguistics (Volume 1: Long Papers)}, pages 66--75,
  Melbourne, Australia. Association for Computational Linguistics.

\bibitem[{Kumar et~al.(2014)Kumar, Kumar, Swathikiran, and
  James}]{kumar2014ancient}
Neethu~S Kumar, Dinesh~S Kumar, S~Swathikiran, and Alex~Pappachen James. 2014.
\newblock Ancient indian document analysis using cognitive memory network.
\newblock In \emph{International Conference on Advances in Computing,
  Communications and Informatics}.

\bibitem[{Lan et~al.(2019)Lan, Chen, Goodman, Gimpel, Sharma, and
  Soricut}]{lan2019albert}
Zhenzhong Lan, Mingda Chen, Sebastian Goodman, Kevin Gimpel, Piyush Sharma, and
  Radu Soricut. 2019.
\newblock Albert: A lite bert for self-supervised learning of language
  representations.
\newblock In \emph{Proc. the International Conference on Learning
  Representations (ICLR)}.

\bibitem[{Lee and Seung(2001)}]{lee2001algorithms}
Daniel~D Lee and H~Sebastian Seung. 2001.
\newblock Algorithms for non-negative matrix factorization.
\newblock In \emph{Proc. the Advances in Neural Information Processing Systems
  (NIPS)}.

\bibitem[{Lee et~al.(2009)Lee, Yang, and Park}]{lee2009orbital}
Ki-Won Lee, Hong-Jin Yang, and Myeong-Gu Park. 2009.
\newblock Orbital elements of comet c/1490 y1 and the quadrantid shower.
\newblock \emph{Monthly Notices of the Royal Astronomical Society}.

\bibitem[{Li et~al.(2018)Li, Xiao, Li, Wang, Xu, and Zhu}]{li-etal-2018-simple}
Yanyang Li, Tong Xiao, Yinqiao Li, Qiang Wang, Changming Xu, and Jingbo Zhu.
  2018.
\newblock \href {https://doi.org/10.18653/v1/P18-2047} {A simple and effective
  approach to coverage-aware neural machine translation}.
\newblock In \emph{Proceedings of the 56th Annual Meeting of the Association
  for Computational Linguistics (Volume 2: Short Papers)}, pages 292--297,
  Melbourne, Australia. Association for Computational Linguistics.

\bibitem[{Lin(2004)}]{lin-2004-rouge}
Chin-Yew Lin. 2004.
\newblock \href {https://www.aclweb.org/anthology/W04-1013} {{ROUGE}: A package
  for automatic evaluation of summaries}.
\newblock In \emph{Text Summarization Branches Out}, pages 74--81, Barcelona,
  Spain. Association for Computational Linguistics.

\bibitem[{Liu et~al.(2019{\natexlab{a}})Liu, Yang, Qu, and Lv}]{liu2019ancient}
Dayiheng Liu, Kexin Yang, Qian Qu, and Jiancheng Lv. 2019{\natexlab{a}}.
\newblock Ancient--modern chinese translation with a new large training
  dataset.
\newblock \emph{ACM Transactions on Asian and Low-Resource Language Information
  Processing}.

\bibitem[{Liu et~al.(2019{\natexlab{b}})Liu, Jiang, He, Chen, Liu, Gao, and
  Han}]{liu2019variance}
Liyuan Liu, Haoming Jiang, Pengcheng He, Weizhu Chen, Xiaodong Liu, Jianfeng
  Gao, and Jiawei Han. 2019{\natexlab{b}}.
\newblock On the variance of the adaptive learning rate and beyond.
\newblock In \emph{Proc. the International Conference on Learning
  Representations (ICLR)}.

\bibitem[{Liu et~al.(2019{\natexlab{c}})Liu, Ott, Goyal, Du, Joshi, Chen, Levy,
  Lewis, Zettlemoyer, and Stoyanov}]{liu2019roberta}
Yinhan Liu, Myle Ott, Naman Goyal, Jingfei Du, Mandar Joshi, Danqi Chen, Omer
  Levy, Mike Lewis, Luke Zettlemoyer, and Veselin Stoyanov. 2019{\natexlab{c}}.
\newblock Roberta: A robustly optimized bert pretraining approach.
\newblock \emph{arXiv:1907.11692}.

\bibitem[{Mimno(2012)}]{mimno2012computational}
David Mimno. 2012.
\newblock Computational historiography: Data mining in a century of classics
  journals.
\newblock \emph{Journal on Computing and Cultural Heritage}.

\bibitem[{Papineni et~al.(2002)Papineni, Roukos, Ward, and
  Zhu}]{papineni2002bleu}
Kishore Papineni, Salim Roukos, Todd Ward, and Wei-Jing Zhu. 2002.
\newblock Bleu: a method for automatic evaluation of machine translation.
\newblock In \emph{Proc. the Annual Meeting of the Association for
  Computational Linguistics (ACL)}.

\bibitem[{Radford et~al.(2019)Radford, Wu, Child, Luan, Amodei, and
  Sutskever}]{radford2019language}
Alec Radford, Jeffrey Wu, Rewon Child, David Luan, Dario Amodei, and Ilya
  Sutskever. 2019.
\newblock Language models are unsupervised multitask learners.
\newblock \emph{OpenAI Blog}.

\bibitem[{Shindell et~al.(2001)Shindell, Schmidt, Mann, Rind, and
  Waple}]{shindell2001solar}
Drew~T Shindell, Gavin~A Schmidt, Michael~E Mann, David Rind, and Anne Waple.
  2001.
\newblock Solar forcing of regional climate change during the maunder minimum.
\newblock \emph{Science}.

\bibitem[{Stephenson and Willis(2008)}]{stephenson2008vapours}
F~Richard Stephenson and David~M Willis. 2008.
\newblock ‘vapours like fire light’are korean aurorae.
\newblock \emph{Astronomy \& Geophysics}.

\bibitem[{Tang et~al.(2018)Tang, Cap, Pettersson, and
  Nivre}]{tang-etal-2018-evaluation}
Gongbo Tang, Fabienne Cap, Eva Pettersson, and Joakim Nivre. 2018.
\newblock \href {https://www.aclweb.org/anthology/C18-1112} {An evaluation of
  neural machine translation models on historical spelling normalization}.
\newblock In \emph{Proceedings of the 27th International Conference on
  Computational Linguistics}, pages 1320--1331, Santa Fe, New Mexico, USA.
  Association for Computational Linguistics.

\bibitem[{Vaswani et~al.(2017)Vaswani, Shazeer, Parmar, Uszkoreit, Jones,
  Gomez, Kaiser, and Polosukhin}]{vaswani2017attention}
Ashish Vaswani, Noam Shazeer, Niki Parmar, Jakob Uszkoreit, Llion Jones,
  Aidan~N Gomez, {\L}ukasz Kaiser, and Illia Polosukhin. 2017.
\newblock Attention is all you need.
\newblock In \emph{Proc. the Advances in Neural Information Processing Systems
  (NIPS)}.

\bibitem[{Yang et~al.(2005)Yang, Park, and Park}]{yang2005analysis}
Hong-Jin Yang, Changbom Park, and Myeong-Gu Park. 2005.
\newblock Analysis of historical meteor and meteor shower records: Korea,
  china, and japan.
\newblock \emph{Icarus}.

\bibitem[{Yang et~al.(2011)Yang, Torget, and Mihalcea}]{yang-etal-2011-topic}
Tze-I Yang, Andrew Torget, and Rada Mihalcea. 2011.
\newblock \href {https://www.aclweb.org/anthology/W11-1513} {Topic modeling on
  historical newspapers}.
\newblock In \emph{Proceedings of the 5th {ACL}-{HLT} Workshop on Language
  Technology for Cultural Heritage, Social Sciences, and Humanities}, pages
  96--104, Portland, OR, USA. Association for Computational Linguistics.

\bibitem[{Yoo et~al.(2015)Yoo, Park, Kim, Choi, Sin, and
  Jun}]{yoo2015classification}
Chulsang Yoo, Minkyu Park, Hyeon~Jun Kim, Juhee Choi, Jiye Sin, and Changhyun
  Jun. 2015.
\newblock Classification and evaluation of the documentary-recorded storm
  events in the annals of the choson dynasty (1392--1910), korea.
\newblock \emph{Journal of Hydrology}.

\bibitem[{You et~al.(2017)You, Gitman, and Ginsburg}]{you2017large}
Yang You, Igor Gitman, and Boris Ginsburg. 2017.
\newblock Large batch training of convolutional networks.
\newblock \emph{arXiv:1708.03888}.

\bibitem[{Zhang et~al.(2014)Zhang, Zhang, Che, and Liu}]{zhang2014character}
Meishan Zhang, Yue Zhang, Wanxiang Che, and Ting Liu. 2014.
\newblock Character-level chinese dependency parsing.
\newblock In \emph{Proc. the Annual Meeting of the Association for
  Computational Linguistics (ACL)}.

\bibitem[{Zhang et~al.(2019{\natexlab{a}})Zhang, Han, Liu, Jiang, Sun, and
  Liu}]{zhang-etal-2019-ernie}
Zhengyan Zhang, Xu~Han, Zhiyuan Liu, Xin Jiang, Maosong Sun, and Qun Liu.
  2019{\natexlab{a}}.
\newblock \href {https://doi.org/10.18653/v1/P19-1139} {{ERNIE}: Enhanced
  language representation with informative entities}.
\newblock In \emph{Proceedings of the 57th Annual Meeting of the Association
  for Computational Linguistics}, pages 1441--1451, Florence, Italy.
  Association for Computational Linguistics.

\bibitem[{Zhang et~al.(2019{\natexlab{b}})Zhang, Li, and
  Su}]{zhang2019automatic}
Zhiyuan Zhang, Wei Li, and Qi~Su. 2019{\natexlab{b}}.
\newblock Automatic translating between ancient chinese and contemporary
  chinese with limited aligned corpora.
\newblock In \emph{CCF International Conference on Natural Language Processing
  and Chinese Computing}.

\bibitem[{Zhang(1985)}]{zhang1985korean}
ZW~Zhang. 1985.
\newblock Korean auroral records of the period ad 1507-1747 and the sar arcs.
\newblock \emph{Journal of the British Astronomical Association}.

\bibitem[{Zhao et~al.(2014)Zhao, Wu, Wang, and Shi}]{zhao2014sentiment}
Huidong Zhao, Bin Wu, Haoyu Wang, and Chuan Shi. 2014.
\newblock Sentiment analysis based on transfer learning for chinese ancient
  literature.
\newblock In \emph{International Conference on Behavioral, Economic, and
  Socio-Cultural Computing}.

\end{thebibliography}
\bibliographystyle{acl_natbib}



\end{document}